\documentclass[12pt]{article}

\usepackage{amssymb}
\usepackage[]{changes}
\usepackage{tabularx}
\usepackage{multirow} 
\usepackage{array}
\usepackage{float}
\usepackage{svg}
\usepackage{geometry}
\usepackage{booktabs}
\usepackage{enumitem}
\usepackage{authblk}

\title{AI-Driven Decision Support in Oncology: \\Evaluating Data Readiness for Skin Cancer Treatment}

\author[1,2]{Joscha Grüger}
\author[2]{Tobias Geyer}
\author[3]{Tobias Brix}
\author[3]{Michael Storck}
\author[4]{Sonja Leson}
\author[4]{Laura Bley}
\author[4]{Carsten Weishaupt}
\author[1,2]{Ralph Bergmann}
\author[4,5]{Stephan A. Braun}

\affil[1]{Business Information Systems II, University of Trier, Behringstraße 21, 54296 Trier, Germany}
\affil[2]{German Research Center for Artificial Intelligence (DFKI), Trier Branch, Behringstraße 21, 54296 Trier, Germany}
\affil[3]{Institute of Medical Informatics, University of Münster, Schweitzer-Campus 1, 48149 Münster, Germany}
\affil[4]{Department of Dermatology, University Hospital Münster, Von-Esmarch-Straße 58, 48149 Münster, Germany}
\affil[5]{Department of Dermatology, Medical Faculty, Heinrich Heine University, Universitätsstr. 1, 40225 Düsseldorf, Germany}

\date{}  

\begin{document}

\maketitle  

\begin{abstract}
This research focuses on evaluating and enhancing data readiness for the development of an Artificial Intelligence (AI)-based Clinical Decision Support System (CDSS) in the context of skin cancer treatment. The study, conducted at the Skin Tumor Center of the University Hospital Münster, delves into the essential role of data quality, availability, and extractability in implementing effective AI applications in oncology. By employing a multifaceted methodology, including literature review, data readiness assessment, and expert workshops, the study addresses the challenges of integrating AI into clinical decision-making. The research identifies crucial data points for skin cancer treatment decisions, evaluates their presence and quality in various information systems, and highlights the difficulties in extracting information from unstructured data. The findings underline the significance of high-quality, accessible data for the success of AI-driven CDSS in medical settings, particularly in the complex field of oncology.
\end{abstract}


\section{Introduction}
\label{introduction}
Over the past few years, the field of artificial intelligence (AI) has shown great promise in various domains, including medicine. A potential use case for AI in medicine is its application in managing advanced-stage cancer treatment, where limited evidence often makes treatment choices reliant on the personal expertise of the physicians. The complex nature of oncological disease processes and the multitude of factors that need to be considered when making treatment decisions make it difficult to rely solely on evidence-based trial data, which is often limited and may exclude certain patient populations. This results in physicians making decisions on a case-by-case basis, drawing on their experience of previous cases, which is not always objective and may be limited by the small number of cases they have observed. 

In this context, the use of clinical decision support systems (CDSS) using similarity-based AI approaches can potentially contribute to better oncology treatment by supporting physicians in the selection of treatment methods \cite{DePaz2009Apr,Sutton2020Feb}. One approach is Case-Based Reasoning (CBR), a subfield of AI that deals with experience-based problem solving. It involves problem solving through the reuse and adaptation of empirical knowledge. In this approach, potentially relevant cases and solutions are searched for in the documented cases with the help of similarity measures based on analogies to a given case. The approach is used in many fields such as medicine, law, and engineering. The relationship to analogy-based clinical reasoning makes CBR a promising approach to analogy-based decision support in medicine \cite{BergmannMBAM20,Choudhury.2016}.

However, despite years of research, medicine is still at the beginning stages of incorporating AI into clinical decision-making. One of the main reasons for this is the lack of high-quality data, concerns around privacy and the need for explainability of results \cite{Kelly2019Dec}. In addition, the lack of structured data and the widespread use of free-text documentation means that there is little understanding of which information is included in the data and which is not. Moreover, the effectiveness of Clinical Decision Support Systems (CDSS) hinges critically on the quality, availability, and timeliness of the data upon which they are based. High-quality data ensures that the insights and recommendations generated by CDSS are reliable and actionable. Equally important is the accessibility of pertinent data, which must be both relevant and current to inform clinical decisions effectively \cite{Sutton2020Feb}.

In this study, we assessed the data availability for a prospective AI-based Clinical Decision Support System (CDSS) aimed at cancer patient treatment. Therefore, we present a structured process to interdisciplinary analyze the AI-readiness of clinical data. We applied this on patients of the scin cancer treatment. To this end, we analyzed the extractability of data, identified clinically relevant data points for analogy-based reasoning in a cohort of selected skin tumor patients, and evaluated the quality of the data. Utilizing the extracted data, we investigated the presence of medically significant data points and evaluated the data's readiness for AI application. Furthermore, we identified opportunities for information extraction and outlined necessary actions for data collection improvement.

\section{Related Work}
\subsection{Data Quality and Clinical Decision Support}
The impact of medical data quality in the context of AI and decision support systems has already been discussed across different aspects in the literature \cite{Sutton2020Feb}.
Since such systems are AI-based nowadays, they generally require high-quality training and testing data in many respects, as Budach et al. \cite{Budach.2022} point out. Conversely, this implies that such application systems relying on poor data quality are ineffective. Despite the time and effort required to prepare a high-quality data foundation, it is essential for training and testing. Neglecting this can result in the identification of false patterns and conclusions, leading to unreliable decision support suggestions.

Berner et al. \cite{Berner.2005} emphasized the critical role of data quality in the development of CDSS. Their study specifically examined how the completeness and accuracy of medical records significantly influenced a CDSS designed for risk assessment for gastrointestinal bleeding and therapy recommendation when prescribing non-steroidal anti-inflammatory drugs. Therefore, medical dictations were analyzed to determine the extent to which the statements were complete and accurate. It was shown that CDSS worked most efficiently when the data in medical records was of high quality. Furthermore, it was demonstrated that these tools cannot reach their full potential unless efforts were made to improve the accuracy and completeness of clinical data in medical records \cite{Berner.2005}.  
 
The inaccuracy and incompleteness of medical data leads with regard to CDSS to the possibility that the medical decisions facilitated by the system may result in negative patient outcomes. While Berner et al. \cite{Berner.2005} shed light on how incomplete and inaccurate medical records can affect CDSS, Hasan and Padman \cite{Hasan.2006} continued by proposing a solution-oriented approach by simulation and regression, quantifying the relative impact of poor data quality on overall CDSS accuracy. The results of this analysis can support the development of procedures to minimize incorrect medical decisions facilitated by these systems \cite{Hasan.2006}. 

Similar to Hasan and Padman \cite{Hasan.2006}, Chen and Ren \cite{Chen.2023} proposed their approach. They stated that most of the current Electronic Medical Records (EMR) data quality evaluation methods were based on some conventional evaluation indicators, and rarely considered the introduction of clinical evidence which is why they proposed an evaluation approach based on clinical evidence and a deep text matching model \cite{Chen.2023}. As a result, the approach was able to effectively discriminate between high-quality and low-quality EMRs, with the high-quality EMRs found generally containing sufficient and consistent information for disease diagnosis \cite{Chen.2023}. 
 

Li et al. \cite{Li.2021} further emphasized the need to establish data quality standards to improve AI-driven tools. Here in particular, the processes around data collection, storage, and annotation were mentioned as significant effects on data quality and application effectiveness and that exactly these areas of clinical data were inconsistent in each hospital \cite{Li.2021}. This led to different levels of quality and to difficulties in the coherence of data collected at different sites, which is why it is crucial for the field to formulate a series of criteria for multidisciplinary clinical data collection, storage, and annotation \cite{Li.2021}. 


The literature has consistently emphasized the central role of data availability and quality in the development of CDSS. This is also underlined by the work of Lewis et al. \cite{Lewis.2023} in which a literature review of approaches and tools for assessing the data quality of EHRs was conducted. To the authors' best knowledge, no study has yet addressed the facilitation of data selection and assessment processes for CDSS, both in a broader context and specifically in the field of oncology.

\subsection{AI Readiness}\label{sec:aiready}
In the landscape of AI readiness research, substantial attention has been devoted to evaluating organizational readiness, often overlooking the crucial data-related dimensions. Within the realm of organizational readiness, Jöhnk et al. \cite{Jhnk2020} conducted an extensive investigation. They engaged in a comprehensive interview-based study involving AI experts, leading to the identification of five foundational categories crucial for organizational AI readiness: strategic alignment, resources, knowledge, culture, and data and 18 factors within these categories. Notably, within the data category, their research delineated the factors data availability, data quality, data accessibility, and data flow \cite{Jhnk2020}. Alsheibani et al. \cite{Alsheibani2018ArtificialIA} focused on the adoption of AI by employing the Technology-Organization-Environment (TOE) framework \cite{Tornatzky1990}. The TOE framework describes a theoretical framework defining the adoption of technology within organizational settings. It delineates how the process of embracing and incorporating technological innovations is shaped and guided by the interplay of the technological landscape, organizational dynamics, and environmental factors \cite{Tornatzky1990}. Pumplun et al. \cite{Pumplun2019ANO} further contributed to this discourse by extending the TOE framework with AI-specific dimensions. Based on the TOE framework and the results of interviews with 12 AI experts, they expanded the TOE framework adapted to the specific requirements of artificial intelligence adoption. Regarding the data aspect, the authors contend that there exist limited and incomplete criteria for initially and evaluating the data \cite{Pumplun2019ANO}.

However, the prevailing focus on the organizational facet of AI readiness often leaves unexplored the intricate technical aspects that are equally essential for successful AI implementation. Recognizing this gap, our research embarks on a distinct path, one that combines insights from the existing literature with a specialized emphasis on data-centric AI-readiness evaluation. We contend that for solving specific problems using AI technology, such as providing decision support for skin cancer treatment, the readiness evaluation must encompass the data capabilities for the specific problem. 

\section{Materials and Methods}

\begin{figure}[H]
    \centering
    \includegraphics[width=0.99\linewidth]{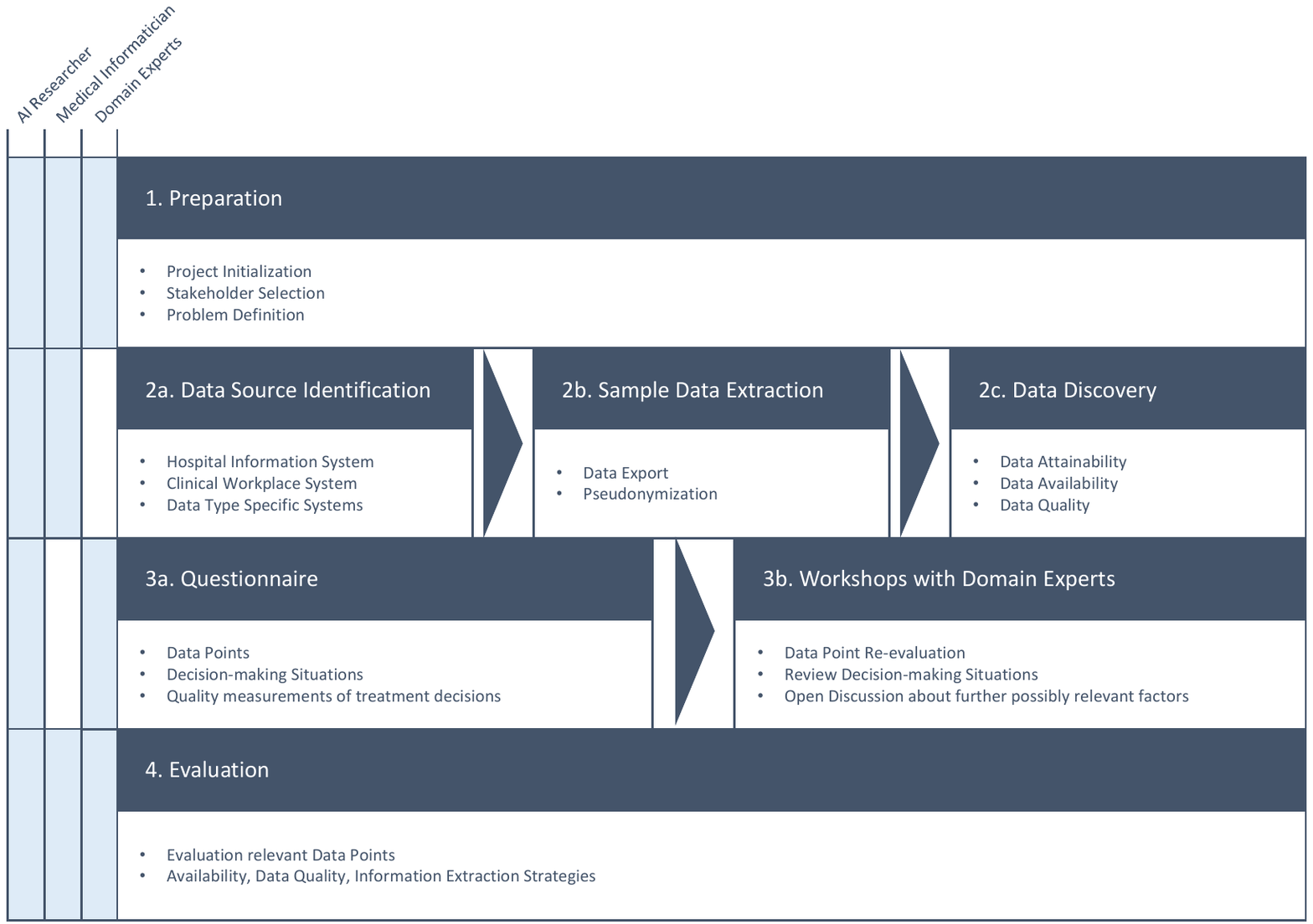}
    \caption{The methodological approach was divided into the following steps: \textit{Preparation} (1.) describes essential tasks that form the basis for the further procedure. \textit{Data Source Identification} (2a.) describes the procedure for determining the correct data sources. \textit{Sampla Data Extraction} (2b.) describes the export and transfer of medical data. \textit{Data Discovery} (2c.) describes the evaluation of data suitability within the context of an AI project. \textit{Questionnaire} (3a.) describes the development of the questionnaire, which enables the collection of decision-relevant data and their prioritization. \textit{Workshops with Domain Experts} (3c.) describes the organization of the workshops in which the results of the questionnaire were discussed. \textit{Evaluation} (4.) describes the examination of the data relevant for a treatment decision.}
    \label{fig:Framework_Overview}
\end{figure}

The approach of this study comprised six steps: \textit{Preparation}, \textit{Data Source Identification}, \textit{Sample Data Extraction}, \textit{Data Discovery}, \textit{Questionnaire}, \textit{Workshops with Domain Experts}, and \textit{Evaluation} (see Figure \ref{fig:Framework_Overview}). This methodology is grounded in the technical and data-specific aspects of several AI-readiness frameworks \cite{Jhnk2020,Alsheibani2018ArtificialIA,Tornatzky1990,Pumplun2019ANO} (see section \ref{sec:aiready}).

During the \textit{Preparation}, stakeholders were identified, selected, and a common understanding of the problem was derived. The goal of the \textit{Data Source Identification} was to identify all data sources relevant for the project, to conduct a preliminary investigation of the data and their extractability, and to make a well-founded system selection in the context of the defined objective. In the \textit{Sample Data Extraction} phase, the concrete extraction of a sample data set was performed. This formed the basis for the subsequent \textit{Data Discovery} and thus the analysis of the data with regard to its usability for the defined AI project. The next step was to evaluate the data from a medical perspective. Divided into a \textit{questionnaire} and two \textit{workshops}, the relevance of the existing data points for the given problem was evaluated and any missing data points were identified. 

\subsection{Preparation}
The Preparation phase consisted of the tasks project initialization, stakeholder selection and problem definition. The goal was to map all required competencies in the team and to establish a common understanding of the problem. The team needed to include competencies of the involved medical domains, the healthcare information systems and the data sources available at the hospitals and competencies in AI and data processing. 

In this study, the feasibility of a case-based CDSS from a data perspective was investigated. Exemplary this was done for the Skin Tumor Center of the UKM. The result of the study is the evaluation of the existing data with regard to the use case and the detection of potentially missing data.  

The medical expertise was represented by four dermato-oncologists with many years of professional experience of the Skin Tumor Center of the UKM. Knowledge of existing data systems, their data structures and export possibilities was contributed by two experts from the Institute of Medical Informatics Münster. The AI perspective was represented by three researchers from the German Institute for Artificial Intelligence (DFKI).  

\subsection{Data Source Identification}

To prepare for the skin cancer use case at the UKM, the focus was on accessible information systems within the skin tumor center. This ensured compliance with data protection and medical confidentiality. The decision was to utilize all data available in the Hospital Information System (HIS), including information from the Laboratory Information System (LIS) and Radiology Information System (RIS). While briefly considering image data from the Picture Archiving and Communication System (PACS), it wasn't the primary focus for this use case.

In addition, data from the state-specific cancer registries was used. In Germany, cancer treatment data is mandatorily collected in state-specific cancer registries \cite{Holleczek2017Sep}. This information is extracted using ONDIS software provided by the Westfalen-Lippe Association of Health Insurance Physicians (KVWL). Although this data mirrors that in the EMR, its structured format is more conducive for AI applications.

\subsection{Sample Data Extraction}
For the export and usage of medical data, obtaining consent and ethical agreement is essential due to the highly private and sensitive nature of patient information. Therefore, the export and sharing of medical data were based on informed patient consent, which was issued specifically for this research project. In this study, 5 patients (2 female, 3 male) consented. The procedure was approved by the local ethics committee of the University of Muenster, Germany\footnote{Ethik-Kommission Westfalen-Lippe: \# 2020-652-f-S; 22 September 2020}. There are two important aspects to note regarding the selection of data for the study. Firstly, care was taken to ensure that the data used were as representative as possible by selecting complex patients (patients with complex treatment histories and many data points). Secondly, the selected patients are not a representative set for a case base of a CDSS, nor is an attempt made to draw a decision-supporting conclusion on the basis of their data.  The patients' data serve as a data basis that is to be evaluated in this study. All medical data from the HIS and ONDIS were exported, pseudonymized, and forwarded to DFKI for analysis. In total, 7 categories of data were exported (see Table \ref{tab:data_description}).
\begin{table}[H]
\centering
\small
\resizebox{\linewidth}{!}{%
\begin{tabular}{|p{0.2\linewidth}|p{0.7\linewidth}|}
\hline
\textbf{Data Category} & \textbf{Description} \\ \hline
ONDIS & Cancer registry data for the 5 patients were exported and each provided as a single XML file including patient demographics, diagnosis, treatment, and outcomes. \\ \hline
Personal Data & The personal and identifying data of the patients such as name, hospital ID, gender, etc. were exported and provided as CSV file. \\ \hline
Cases & Meta information for all hospital cases of the 5 patients at the UKM, provided as CSV file. There were 171 cases in total, with a distribution between 20–59 cases per patient. The data included the date of admission, type of stay, etc. \\ \hline
Diagnoses & All 367 diagnoses stored in the HIS were exported for the 171 cases. In addition to the International Classification of Diseases (ICD) codes, information was also provided on the localization and whether the diagnoses were primary or secondary diagnoses and so on. \\ \hline
Procedures & All 288 procedures stored in the HIS were exported for the 171 cases. In addition to the Operation and Procedures Key (OPS) codes, the date of the procedure and other information was also exported. \\ \hline
Laboratory values & 7656 laboratory values distributed over 247 laboratory findings were exported. In addition to the type of laboratory value, normal ranges, units, etc. were also exported. \\ \hline
Medical data & This was the largest and at the same time most unstructured part of the data export. In the HIS, medical information was stored as forms with individual input fields. Thus, the former paper-based patient record was mapped more or less one-to-one electronically. A total of 43,570 forms with a total of 2,209,463 data fields were exported for the 5 patients, resulting in an average of 8,714 forms and 441,892.6 data fields per patient. The data types were heterogeneous, ranging from numbers to selection fields to long free texts. \\ \hline
\end{tabular}%
}
\caption{Description of Data Export}
\label{tab:data_description}
\end{table}

Prior to disclosure, as required by the consent, the plain names, addresses, and public identifiers such as insurance numbers were replaced with speaking pseudonyms. For example, ``Max Miller" $\rightarrow$ ``John Doe". In this way, the risk of re-identification was reduced to a required minimum, but the reading flow in the texts for the AI was not impaired. 

\subsection{Data Discovery}\label{sec:datadiscovery}
The data discovery phase encompassed an evaluation of data suitability within the context of an AI project. During this phase, an in-depth analysis of the data was conducted regarding the specified objective or use case. In the present project, the use case centered on the creation of a CDSS tailored for the treatment of patients with skin cancer. The underlying quality criteria were collected as part of a literature study, aggregated and selected by the goal of use to determine AI readiness, and divided into two categories (see Table \ref{tab:dqi}) \cite{Jhnk2020, Strong2002Aug,Wang2015Dec,Pipino2003Jul,Cai2015May,Batini2016Jun}.
\begin{table}[htbp]
    \centering
  
    \begin{tabular}{|p{0.3\linewidth}|p{0.7\linewidth}|}
        \hline
        \textbf{Aspect} & \textbf{Description} \\
        \hline
        Data attainability and availability & \begin{itemize}[leftmargin=*]
                                                \item \textbf{Currency (timeliness):} Data should be updated regularly for use within an AI system. It is important that the data is current enough to be meaningful for the context and AI approach used.
                                                \item \textbf{Amount of Data (Data Availability):} The Availability criterion refers to the amount of data that is appropriate and sufficient for the task at hand.
                                             \end{itemize} \\
        \hline
        Data quality & \begin{itemize}[leftmargin=*]
                           \item \textbf{Accuracy/Free-of-Error:} Accuracy is the degree of correctness and precision with which information represents certain states of the real world \cite{Cai2015May}.
                           \item \textbf{Trust (Trustworthiness):} Trustworthiness is used to evaluate non-numeric data and refers to whether a piece of information can be considered true and trustworthy.
                           \item \textbf{Completeness:} Completeness is defined as the degree to which a given data set contains data and attributes that describe the corresponding set of real-world objects.
                           \item \textbf{Consistency:} Data consistency refers to whether logical relationships between correlated data are correct and complete. This means that the same data, but stored in different sources, should be considered equivalent.
                           \item \textbf{Structure:} Differently structured data can have an impact on AI readiness. For example, structured data stored in (two-dimensional) relational database tables is easier for standardized AI applications to read than unstructured data \cite{Jhnk2020}.
                      \end{itemize} \\
        \hline
    \end{tabular}
      \caption{Data Quality and Availability}\label{tab:dqi}
\end{table}

The data delivery was divided into structured data from the HIS data model, data from so-called forms of the HIS and data from the cancer registry export of the hospital. The structured data from the HIS comprised data on procedures, diagnosis, laboratory and case, divided into 526 columns. The form data originated from HIS user-defined forms, including forms for radiological findings, operation room reports, medication reports, and histology reports. The forms consisted partly of structured data and partly of free text fields. The cancer registry data was based on the ADT/GEKID. The ADT/GEKID basic data set, available at \footnote{https://www.gekid.de/adt-gekid-basisdatensatz}, follows a standardized format for oncological documentation. 

The evaluation of accessibility and availability of data is composed of timeliness and the amount of data available. From a medical perspective and considering the short development cycles for new medications and treatments \cite{Densen2011}, patients within the last 5 years are considered relevant for decision support. Annually, the Skin Tumor Center at the UKM records round about 220 new cases of patients with malignant melanoma. This amount is already sufficient to achieve results with an analogy-based decision support system.  

The first criterion of data quality assessment is accuracy. The accuracy of the data could not be conclusively assessed, as there was no gold standard against which to check for errors or missing data. However, errors such as neologisms and misspellings could be identified in the free texts. The same was true for trustworthiness and completeness, these could not be checked against the data. For automatically collected values, it could be assumed that they were recorded completely, but in the case of free texts, it could not be ruled out that information was forgotten. The consistency of the data could be partially checked and confirmed by comparing the data from the HIS and the cancer registry exports. Inconsistencies existed, however, in the free text fields; these related to the use of different identifiers, spellings, languages and abbreviations for terms, which would make information extraction enormously difficult.

The data of the system tables Procedure, Laboratory and Case were mostly structured. However, they contained two free text fields holding unstructured information. The forms table, on the other hand, contained largely free text. Forms such as the physician's letter, the histology report or the medication report consisted of free text, the information of which is therefore difficult to access and, on the other hand, difficult to anonymize. The situation was different with the cancer registry data, which were completely structured and were entered by a coding specialist.  

The result of the analysis step was a detailed listing and description of all available data as well as an analysis of the data quality. 

\subsection{Questionnaire}
A questionnaire (see \ref{sec:appa}) was developed to collect decision-relevant data and prioritize them. Decision relevance referred to the relevance for a potential decision support system and analogy-based clinical reasoning. The questionnaire was divided into the four subsections: 
\begin{itemize}
    \item     Relevant data points. 

   \item  Data points not previously included (because not collected), but potentially relevant. 

    \item Decision-making situations 

    \item Measures of the quality of treatment decisions 
\end{itemize}
In the first step, physicians listed the 20 most decision-relevant patient data points, then rated their relevance from 1 to 10. In the second question, they identified potentially relevant data not previously considered. In the third question, the participants highlighted several critical decision-making situations in the treatment process where additional support is most needed. The fourth question focused on the measurement of the quality of the treatment decisions. The questionnaire was filled out by three board certified dermatologists specialized in oncology. The results served as an input for the following workshops.

\subsection{Workshops with Domain Experts }
In two workshops with board certified dermatologists specialized in oncology, the results of the questionnaire were discussed in moderated semi-structured group interviews. For this purpose, the results of the questionnaire were aggregated and grouped into thematic areas. 

The workshop's first part focused on decision-relevant data. The questionnaire results were discussed and items were re-evaluated, leading to a consensus-based list. The second part reviewed decision-making situations from the questionnaire, identifying specific scenarios relevant to the use case and assessing their importance for decision support.  Next, endpoints and metrics for treatment effectiveness, such as survival rates or symptom improvement, were discussed and ranked, crucial for evaluating the CDSS. Finally, there was an open discussion on dimensions with limited input into treatment decisions but considered potentially relevant.

\subsection{Evaluation} 
In the evaluation, the data relevant for a treatment decision were examined. This involved evaluating whether the data judged to be relevant was collected and available in one or more data sources, if so, what their data quality was, and how to extract the information. This process included a thorough review of data, based on the data quality criteria (\ref{sec:datadiscovery}).

\section{Results}\label{sec:results}
Regarding the decision-relevant data, a total of 41 explicitly relevant data points were found. Two categories of decision-influencing values were identified. On the one hand, there were medically relevant indicators for the best possible treatment, and on the other hand, there were social and personal factors that influence a treatment decision. Data points relevant from a medical perspective included, e.g., biomarkers, tumor status, possible metastases, or tumor dynamics. From a social perspective, e.g., mobility, place of residence and willingness to travel (for possible study inclusion), subjective quality of life and a desire to have children were relevant. In this study, the expected efficacy of a treatment is based primarily on the statistical results of clinical studies.

\begin{table}
    \centering
    \begin{tabularx}{\textwidth}{|X|c|}
    \hline
    \textbf{Attribute} & \textbf{Relevance} \\
    \hline
         Tumor Stage & \multirow{2}{*}{10} \\
         Expected efficacy of treatment & \\
    \hline 
         Age & \multirow{8}{*}{8} \\ 
         Therapy availability/expertise & \\ 
         Prior therapies (side effects, response) & \\ 
         Patient preference for therapy & \\ 
         Metastases number/size (tumor burden) & \\ 
         Metastasis location & \\ 
         Tumor marker S100, LDH & \\ 
         Mutation status (\textit{melanoma relevant mutations such as BRAF V600, NRAS and CKIT mutations}) & \\ 
         Concomitant diseases & \\
    \hline
         Study availability & \multirow{4}{*}{7} \\ 
         Disease dynamics & \\ 
         Symptomatology of tumor manifestation & \\ 
         General condition (ECOG) & \\
    \hline
         Family planning & 6 \\
    \hline
         Therapy burden/effort & \multirow{3}{*}{5} \\ 
         Melanoma type: mucosal, uveal, cutaneous & \\ 
         Concomitant medication & \\
    \hline
         Compliance & 4 \\
    \hline
    \end{tabularx}
    \caption{Consensus-based Ranking of Decision-Relevant Data Points and Comparative Relevance in Treatment Decisions. 10 for strong relevance and 1 for irrelevant.}
    \label{tab:relevance}
\end{table}

The discussion resulted in a consensus-based list of 20 crucial data points with their relevance assessments (Table \ref{tab:relevance}).  Concerning the decision-making situations, the participants highlighted several critical decision-making situations in the treatment process where additional support is most needed (see Table \ref{tab:situations}). Participants identified critical decision-making situations in treatment, notably managing tumor progression and therapy termination decisions, both rated with a relevance score of 10. Equally vital was the management of side effects, which scored a 9 in relevance. Stage IV tumor progression received a relevance score of 8, highlighting decision-making complexity in advanced disease stages. These scenarios underscore the need for support in clinical decision-making, considering treatment effectiveness, patient safety, and cancer progression dynamics (Table \ref{tab:situations})."

\begin{table}
    \centering    
\begin{tabular}{|m{11cm}|c|}
\hline
\textbf{Situation} & \textbf{Relevance} \\
\hline
Tumor progression under adjuvant therapy & 10 \\
Deciding whether therapy can be terminated & 10 \\
Management of side effects when they occur & 9 \\
Tumor progression in Stage IV & 8 \\
Ordering of therapy options when multiple treatments are available & 8 \\
Reinitiating therapies after the occurrence of side effects & 8 \\
Decision on whether to administer adjuvant therapy & 6 \\
Decision-making at stage change (e.g., II to III or III to IV) & 5 \\
Decision-making at initial diagnosis & 3 \\
\hline
\end{tabular}
\caption{Consensus-based Priority Ranking of Decision-Making Situations in Skin Cancer Treatment}
    \label{tab:situations}
\end{table}

In addition, it was found in this step that, depending on the decision-making situation and the patient's progress in treatment, the accumulated treatment experience about a patient and knowledge about individual response to therapies or side effects play a major role. Situations also differ in when they occur. Examples here are: Is there progression during adjuvant therapy or after adjuvant therapy? Situations also differ in when they occur. For example, whether a progression occurs during adjuvant therapy or after adjuvant therapy. Decision-making processes were not regularly documented. There was not always a documented rationale for decisions, so the documentation did not always make it possible to understand why a particular decision was made. 

Furthermore, the following endpoints, currently relevant to domain experts, were identified and ranked as shown:  
\begin{itemize}
    \item 

    Progression-Free Survival (PFS)

       \item  Overall Survival (OS)

       \item  Quality of Life (QoL) of Patients

      \item   Response rates
\begin{itemize}
  
     \item    Partial Response (PR)

    \item     Complete Response (CR)

\end{itemize}
       \item  Disease Control Rate (DCR)

       \item  Durable response rate (DRR)

       \item  Time to next treatment (TTNT)
\end{itemize}

There was an open discussion of dimensions that have had little to no input into treatment decisions to date but are considered possibly relevant. The domain experts had a consensus that the procedural perspective on the data was likely to be relevant to treatment. Among other things, they included the sequence of therapies, the spacing of therapies, and combinations of therapies. These factors are likely to have an impact on the course of treatment, but so far are unlikely to be incorporated into treatment decision-making in hospitals because of their complexity. 

Finally, regarding the data availability, of the 41 data points judged to be relevant, 13 could be extracted from cancer registry exports. These were available in structured form and were therefore easily accessible. Of the data points not included in the cancer registry exports, 20 were contained in the HIS data. Of these, 5 values were available in structured form, including laboratory values and medication. The remaining 15 data points in the HIS were available in unstructured form, in texts written in prose, such as doctors' letters, findings or medical history forms. Extracting information from the free texts would be a non-trivial challenge, as the terminology used was very heterogeneous; German and English abbreviations (e.g., BCC and BZK for Basal Cell Carcinoma), Latin names (e.g., Carcinoma basocellulare / Basal Cell Carcinoma) or neologisms (e.g., Immuno-oncology) were used, and spelling errors occurred. In addition, the texts also contained negated statements that described, for example, what did not apply to a patient. These factors render the extraction of information from unstructured data challenging. Techniques such as natural language processing (NLP) \cite{LeGlaz2021May} and large language models (LLMs) \cite{Dunn2022Dec,Gu2023Jul} could be employed to address this issue, constantly facing the challenge of performing reliable and accurate information extraction, even on complex medical texts. However, the application of these techniques is a major challenge in itself and requires a great deal of effort to implement, as extensive training data is required and the risk of bias must be taken into account and minimized. Of the data points classified as relevant, 8 could not be found in the data of any of the information systems. These included social factors such as quality of life, willingness to travel, a desire to have children, and living situation. In addition, other as yet undocumented data included the organ-specific tumor progression, which describes how a tumor grows and spreads within a particular organ, adapting to and exploiting the unique biological and microenvironmental features of that organ, the occurrence of residual tumors, the rate of progression and the dynamics of the tumor. These values were not yet recorded or not recorded as a fixed part of the documentation but were considered relevant for decision-making by the domain experts. The situation is similar with the expected efficacy of treatment; functional drug screening data is not yet routinely available in most clinics \cite{Dolgin2024Feb}. The first steps towards precision cancer therapy are currently being taken through the increased use of molecular tumor boards. Of course, future CDSS should also include other precision medicine data sets. However, we have not yet reached this point. These data examples represent a significant gap in current data collection and require special attention in terms of future feasible solutions. An important first step is standardized documentation by medical staff. Due to the general overload of medical staff, it is particularly important to work on practicable solutions for time-efficient, structured data entry. AI-based documentation assistants could play a key role here and support doctors in the structured entry of important data points.

Moving forward, the integration of different data types into a coherent, AI-powered clinical decision support system remains a major challenge. While this work identifies a broad range of decision-relevant data points, including medically relevant indicators and social/personal factors, a clear strategy for their integration is to be defined. This gap underscores the ongoing need for research and development efforts aimed at overcoming the practical and feasibility concerns associated with implementing such complex systems in clinical practice.

\section{Summary and conclusions}
The study conclusively demonstrates the imperative role of data quality and availability in the successful implementation of AI-based CDSS in oncology, specifically for skin cancer treatment. The extensive evaluation of data at the Skin Tumor Center of the University Hospital Münster reveals a mix of structured and unstructured data, with significant challenges in extracting relevant information from free-text documentation. The research identifies 20 key data points essential for decision-making in skin cancer treatment. The study emphasizes the need for improved data collection practices, including the integration and documentation of social and personal factors, such as quality of life, willingness to travel or a desire to have children and detailed medical data, to enhance the effectiveness of AI in clinical settings. The findings advocate for a structured approach in medical data documentation and the incorporation of comprehensive data quality standards, also for medical reports and doctor's letters to realize the full potential of AI in healthcare.

In future work, expanding the scope of this study is vital. Integrating a larger and more diverse pool of physicians will enrich the data and insights, making the list of relevant data points more representative of the broader medical community's needs. Concurrently, developing structured documentation standards is imperative for uniform data collection and processing, enhancing AI effectiveness and enabling development of CDSS. Additionally, future research will focus on determining the optimal number of patients required to develop an analogy-based Clinical Decision Support System (CDSS) aimed at skin cancer treatment.

\vspace{1cm}
\vspace{0.5cm}

\noindent\textbf{Funding:} This research was funded by "Nationale Versorgungskonferenz Hautkrebs (NVKH) e.V." grand acronym Pre-Oncocase. We acknowledge support from the Open Access Publication Fund of the University Trier.
\vspace{0.5cm}

\noindent\textbf{Informed Consent Statement:} Written informed consent has been obtained from the patient(s) to publish this paper.
\vspace{0.5cm}

\appendix

\section{Questionnaire}
\label{sec:appa}
\noindent\textit{This questionnaire was translated from German into English.}
\newline
\newline
The aim of the questionnaire is to collect and prioritize decision-relevant data.

\begin{itemize}
\item Decision relevance/relevance refers to the relevance for a decision support system. Is this data that you (would) include in clinical reasoning or not? Is this data that you consider relevant for making treatment decisions in a dermatologic-oncologic context? Could the data be used to compare patients in order to identify similar patients?

\item Data refers to certain real attributes such as: LDH value, age, gender or values/attribute characteristics that are not explicitly recorded
\end{itemize}

\subsection*{Question 1: Decision-relevant data}
\begin{itemize}
\item[1.1] Name the 20 most decision-relevant data of a patient/case that you take into account when making a treatment decision.
\item[1.2] Rate the relevance of the data with points between 1 and 10, with 10 for particularly relevant and 1 for probably less relevant.
\end{itemize}

\subsection*{Question 2: Inclusion of previously unavailable data points}
\begin{itemize}
\item[2.1] Are there data/characteristics of patients that you have not yet included (e.g. because they are not available or because the analysis would be too time-consuming), but which you consider to be potentially relevant?
\item[2.2] Rate the relevance of the data with points between 1 and 10, with 10 for particularly relevant and 1 for probably less relevant.
\end{itemize}

\subsection*{Question 3: Decision-making situations}
Various decision-making situations occur during the treatment process. These can differ, among other things, in the time of occurrence (treatment decision for the treatment of a primary tumor or treatment decision for the occurrence of a recurrence) or with regard to the decision to be made (selection of a treatment method, dosage of a drug, determination of treatment intervals, ...).
\begin{itemize}
\item[3.1] List below the decision-making situations in the course of a patient history of patients in oncological dermatology.
\item[3.2] Are there situations in which you see a particular need for decision support? Rate the need from 1-10, with 10 indicating a particularly high need.
\item[3.3] In the situations mentioned, is there any data that you would include in the decision that you did not mention in the generic description? How do the situations mentioned differ?  
\end{itemize}

\subsection*{Question 4: Measurands for the quality of treatment decisions}
\begin{itemize}
    \item[4.1] The success of treatments and decisions is usually measured by the successful recovery of patients. However, other factors could also be included (e.g. progression-free survival, ...). List the potentially relevant parameters.
    \item[4.2] Rate the potential relevance of the metrics with points between 1 and 10, with 10 for potentially very relevant and 1 for probably not very relevant.

\end{itemize}

 \bibliographystyle{elsarticle-num} 
 \bibliography{main}





\end{document}